\documentclass{article}

\usepackage[colorlinks]{hyperref}       
\usepackage{amsfonts}       
\usepackage{graphicx}
\usepackage{multirow}
\usepackage{multicol}
\usepackage{subcaption}
\usepackage{arxiv}
\usepackage{euscript}
\usepackage{amsmath}

\title{UGCANet: A Unified Global Context-Aware Transformer-based Network with Feature Alignment for Endoscopic Image Analysis}

\author{{Pham Vu Hung} \\
School of Information and Communication Technology\\
Hanoi University of Science and Technology\\
\texttt{hung.pv194069@sis.hust.edu.vn} \\
\And
{Nguyen Duy Manh} \\
School of Information and Communication Technology\\
Hanoi University of Science and Technology\\
\texttt{manh.nd202657m@sis.hust.edu.vn} \\
\And
{Nguyen Thi Oanh} \\
School of Information and Communication Technology\\
Hanoi University of Science and Technology\\
\texttt{oanhnt@soict.hust.edu.vn} \\
\And
{Nguyen Thi Thuy} \\
School of Science, Engineering and Technology\\
RMIT University\\
\texttt{thuy.nguyen43@rmit.edu.vn} \\
\And
{Dinh Viet Sang} \thanks{Corresponding author}\\
School of Information and Communication Technology\\
Hanoi University of Science and Technology\\
\texttt{sangdv@soict.hust.edu.vn} \\
}

\renewcommand{\shorttitle}{UGCANet: An Efficient Transformer based Method for Colon Polyp Segmentation}

\begin{document}
\maketitle

\begin{abstract}
Gastrointestinal endoscopy is a medical procedure that utilizes a flexible tube equipped with a camera and other instruments to examine the digestive tract. This minimally invasive technique allows for diagnosing and managing various gastrointestinal conditions, including inflammatory bowel disease, gastrointestinal bleeding, and colon cancer.
The early detection and identification of lesions in the upper gastrointestinal tract and the identification of malignant polyps that may pose a risk of cancer development are critical components of gastrointestinal endoscopy's diagnostic and therapeutic applications. Therefore, enhancing the detection rates of gastrointestinal disorders can significantly improve a patient's prognosis by increasing the likelihood of timely medical intervention, which may prolong the patient's lifespan and improve overall health outcomes.
This paper presents a novel Transformer-based deep neural network designed to perform multiple tasks simultaneously, thereby enabling accurate identification of both upper gastrointestinal tract lesions and colon polyps.
Our approach proposes a unique global context-aware module and leverages the powerful MiT backbone, along with a feature alignment block, to enhance the network's representation capability. This novel design leads to a significant improvement in performance across various endoscopic diagnosis tasks. Extensive experiments demonstrate the superior performance of our method compared to other state-of-the-art approaches.

\end{abstract}








\section{Introduction}
Digestive tract diseases, particularly colorectal cancer, pose a significant threat to human health \cite{siegel2018cancer}. Accurate diagnosis and classification of these diseases are critical for effective treatment and improvement of patient outcomes \cite{liu2017automatic}. Despite the increasing incidence of these diseases, manual analysis by expert physicians is still the gold standard for diagnosis and classification. However, this method is time-consuming and subject to human error \cite{qin2017computer}.

Recent advancements in computer-aided diagnosis (CAD) have demonstrated considerable potential in enhancing the precision of disease classification \cite{Li2018}. A notable technique in this regard is polyp segmentation, which aims to accurately identify and isolate lesions in images of the digestive tract \cite{Wong2017}. The extracted information can subsequently aid in diagnosing and classifying digestive tract diseases, including potential precursors to cancer. It is imperative to examine the stomach and its relevant structures thoroughly. These cutting-edge advancements are poised to revolutionize disease detection and diagnosis in the medical field.

Accurate anatomical site classification is a crucial aspect of medical image analysis \cite{Girshick2017, Kooi2017, He2017}, enabling the identification of specific structures within the human body. This information is essential for precise diagnosis and effective treatment planning, providing invaluable insights into areas of anomalies or lesions. Recent advancements in deep learning techniques have contributed significantly to the field of anatomical site classification \cite{Girshick2017, Kooi2017, He2017}. Nonetheless, challenges persist due to the variability of imaging modalities and the inherent complexity of the human anatomy. Overcoming these hurdles is vital for advancing medical image analysis and improving patient outcomes.

Accurate classification of lesions and Helicobacter pylori (HP) has garnered increasing attention in the medical field. Lesion classification is crucial in improving diagnoses and treatments for various medical conditions. Image segmentation is sometimes necessary to understand the lesion or polyp location clearly. Polyp segmentation is particularly valuable in the early detection of digestive tract diseases. Manual analysis of medical images is prone to human error and time-consuming, highlighting the need for automated methods. Our study focuses on classification and segmentation, with findings indicating their equal significance and mutual complementarity.

In this paper, we propose a new model that can well solve two tasks at the same time. The experimental results demonstrate superior performance of our model in both single-task and multi-task learning scenarios.
Our main contributions are:

\begin{itemize}
    \item We identified that the existing MiT backbone needed to adequately leverage the channel attention mechanism, resulting in the loss of context information in deeper layers. To overcome this, we incorporated the CGNL module with channel attention in groups to establish local relationships between channel groups.
    \item However, relationships between groups were still lacking, prompting us to add the SE module to bridge the gap between the groups. The combination of these two modules enhanced the channel characteristics.
    \item By combining these two contributions, we present UGCANet, which can extensively leverage the channel attention mechanism and propagate through the FaPN decoder \cite{FaPN}, resulting in segmented output. Additionally, the Fully-connected layer processes some of the feature information to address the classification task.
\end{itemize}

The remaining sections of the paper are structured as follows. We briefly review related work in Section \ref{sec:related}. Then, we present our proposed method in Section \ref{sec:method}. In Section \ref{sec:Experiments}, we showcase the conducted experiments and discuss results. Finally, we conclude the paper and discuss potential future directions in Section \ref{sec:conclusion}.

\section{Related work}
\label{sec:related}
\subsection{Deep learning and medical image segmentation}
\textbf{Deep learning}. Deep learning has been widely used in various fields, such as computer vision, natural language processing, and speech recognition. In recent years, deep learning has made significant progress in these fields and has been applied to various tasks, including image classification, object detection, and segmentation. In the field of image classification, Convolutional Neural Networks (CNNs) have proven to be highly effective in achieving impressive results on large-scale image classification datasets, such as ImageNet \cite{deng2009imagenet}, CIFAR-100 \cite{cifar100}. This highlights the significance of these models as a benchmark for deep learning and image classification research. The advancements in CNNs have paved the way for continued progress in both fields and serve as a foundational foundation for future research \cite{Densenet, Resnet}. Densenet \cite{Densenet} uses dense blocks, where each layer receives inputs from all previous layers in the block. This helps reduce the vanishing gradient problem and makes it easier for the network to propagate information through the network. In ResNet \cite{Resnet}, the network learns residual connections instead of trying to directly learn the desired mapping from inputs to outputs. Subsequently, in the ensuing years, there have been continued advancements in image classification and deep learning. A new method for scaling CNNs that is more efficient and effective than traditional methods. The method, called EfficientNet, balances the trade-off between accuracy and computational cost by scaling the network's depth, width, and resolution in a consistent manner. The authors show that EfficientNet outperforms previous state-of-the-art models on several benchmark datasets for image classification \cite{efficientnet}. The progression has not stopped at that point. A breakthrough as the first transformers architecture appeared when the attention mechanism was first introduced by \cite{attentionisall}. The introduction of the attention mechanism had a significant impact on the thinking of computer vision researchers, leading to the utilization of the self-attention mechanism in computer vision and the emergence of transformer-based proposals for machine vision. 

The Vision Transformer (ViT) demonstrates remarkable performance compared to cutting-edge CNNs and requires significantly fewer computational resources for training \cite{vit}. Unlike traditional CNNs, In ViT \cite{vit}, the input image is first divided into a sequence of non-overlapping patches, which are then treated as tokens. Which
allows the network to focus on different regions of the input image when
making predictions. While ViT is effective for image classification, adapting it for dense predictions at the pixel level, like object detection and segmentation, presents a significant challenge. The Pyramid Vision Transformer (PVT) \cite{pvt} is introduced as a solution to that issue through the creation of a shrinking pyramid and the implementation of spatial-reduction attention (SRA).
Focal \cite{Focal} used a new mechanism called Focal Self-attention that incorporates both fine-grained local and coarse-grained global interactions, CaiT emphasized the significance of utilizing Layer-scale to scale up the depth dimension \cite{CaiT}, LeViT \cite{levit} astutely applied a combination of CNNs and Transformers to create a novel hybrid neural network, resulting in both precise and rapid inference. While ViT and PVT can be limited by distant dependent information DAT \cite{DAT} introduced deformable self-attention module select key-value pairs depending on the data.

\textbf{Medical image segmentaton}. Unet \cite{Unet} has gained popularity in medical image segmentation. The model is designed to handle tasks with multiple scales and uses a combination of downsampling and upsampling layers to maintain high resolution in the segmented output. The key feature of Unet is its symmetrical architecture, which allows for precise localization and preservation of fine details in the segmented image. However, it has a few limitations. One of the main limitations is that it may struggle with small or fine structures, such as tiny vessels or tumors, in medical images.
Additionally, the symmetrical architecture of Unet can lead to a lack of context information and decreased performance in complex or large-scale segmentation tasks. In order to address these limitations, Unet++ \cite{Unet++} and DoubleUnet \cite{DoubleUnet} were proposed. Unet++ introduces a multi-scale learning mechanism, which helps capture contextual information from different scales. This leads to improved performance for small and fine structures in medical images. DoubleUnet, on the other hand, uses two parallel Unets, each with different levels of abstraction and context, to make predictions. This dual-stream approach helps the model capture local and global context information, improving performance in complex and large-scale segmentation tasks. Several models are based on Unet and Transformers, like TransUnet \cite{transunet} and TransFuse \cite{transfuse}. The Hybrid ViT component of TransUNet stacks the CNN and Transformer together, leading to high computational costs. However, this also improves accuracy and performance compared to using either model individually. TransFuse is designed to address the high computational costs of TransUNet. TransFuse utilizes a parallel architecture, which allows it to reduce the computational overhead of the Transformer and CNN components of the network. In TransFuse, the Transformer and CNN components are trained in parallel, allowing faster and more efficient training. The parallel architecture also allows for more efficient use of hardware resources, such as GPUs, which can result in faster and more efficient inference.

\subsection{Lesion Segmentation for endoscopy}

Gastrointestinal (GI) lesion segmentation is crucial in early detection and diagnosis of digestive tract diseases such as esophageal cancer, duodenal ulcer, and colorectal cancer. In recent years, many deep learning-related works have been published in this area. CNN models such as Unet \cite{Unet} or Unet++ \cite{Unet++} utilize skip connections to mitigate information loss caused by stacking numerous convolutional layers. The Unet architecture enhances the overall visibility of the model and enables the integration of multi-scale information. In 2022, Manh et al. \cite{manh2022endounet} proposed an Unet-based multi-tasking model called EndoUnet to simultaneously solve multiple upper GI tasks. Supplementary modules are also a way to enrich information representation, ASPP \cite{ASPP} expanded the field of view to a larger perceptual area, making it suitable for capturing broader contexts. PraNet \cite{pranet} adds an RFB module \cite{RFB} helps generate features with the different receptive fields.
Another approach to enhance the representation of information is utilizing the attention module. EncNet \cite{EncNet}, DFN \cite{DFN} use attention in the feature map's channel dimension to consider the global context, such as the occurrence of different classes in context. CCBANet \cite{ccbanet} proposed an Attention Balance module (BAM). BAM uses an attention mechanism for three distinct regions: the background, polyp, and boundary. BAM intensifies local context information when retrieving features from the encoder block.

\subsection{Image classification for endoscopy}
Although GI endoscopic image documentation is an economical and effective solution for endoscopic reporting, implementing computer-assisted techniques for quality control poses a challenge due to the resemblance of appearance between different anatomical sites and the considerable and inconsistent variability in site appearance across patients.

In 2018, a GoogLeNet-based diagnostic program \cite{takiyama2018automatic} was developed to recognize the anatomical site from 27335 endoscopic images of four major categories (larynx, esophagus, stomach, and duodenum). The CNN is also utilized in disease detection; Lin et al. \cite{lin2019helicobacter} modified Inception V3 to recognize Helicobacter Pylori infection and obtained relatively satisfactory results (accuracy of 95\%). In 2019, a system called WISENSE \cite{wu2019randomised}, which is based on deep learning and reinforcement learning, was proposed by Wu et al. for real-time blind-spot monitoring, timing the procedure, and classifying anatomical sites. In this study, the authors implemented the 27-class protocol, which includes 26 anatomical sites and a NA class for images that cannot be classified to any site. By combining data collection, automatic ROI extraction, and a CNN-based model, He et al. \cite{he2020deep} presented a comprehensive workflow for EGD image classification in 2020.

\section{Method}
\label{sec:method}
\begin{figure*}[ht!]
\centerline{\includegraphics[width=\textwidth]{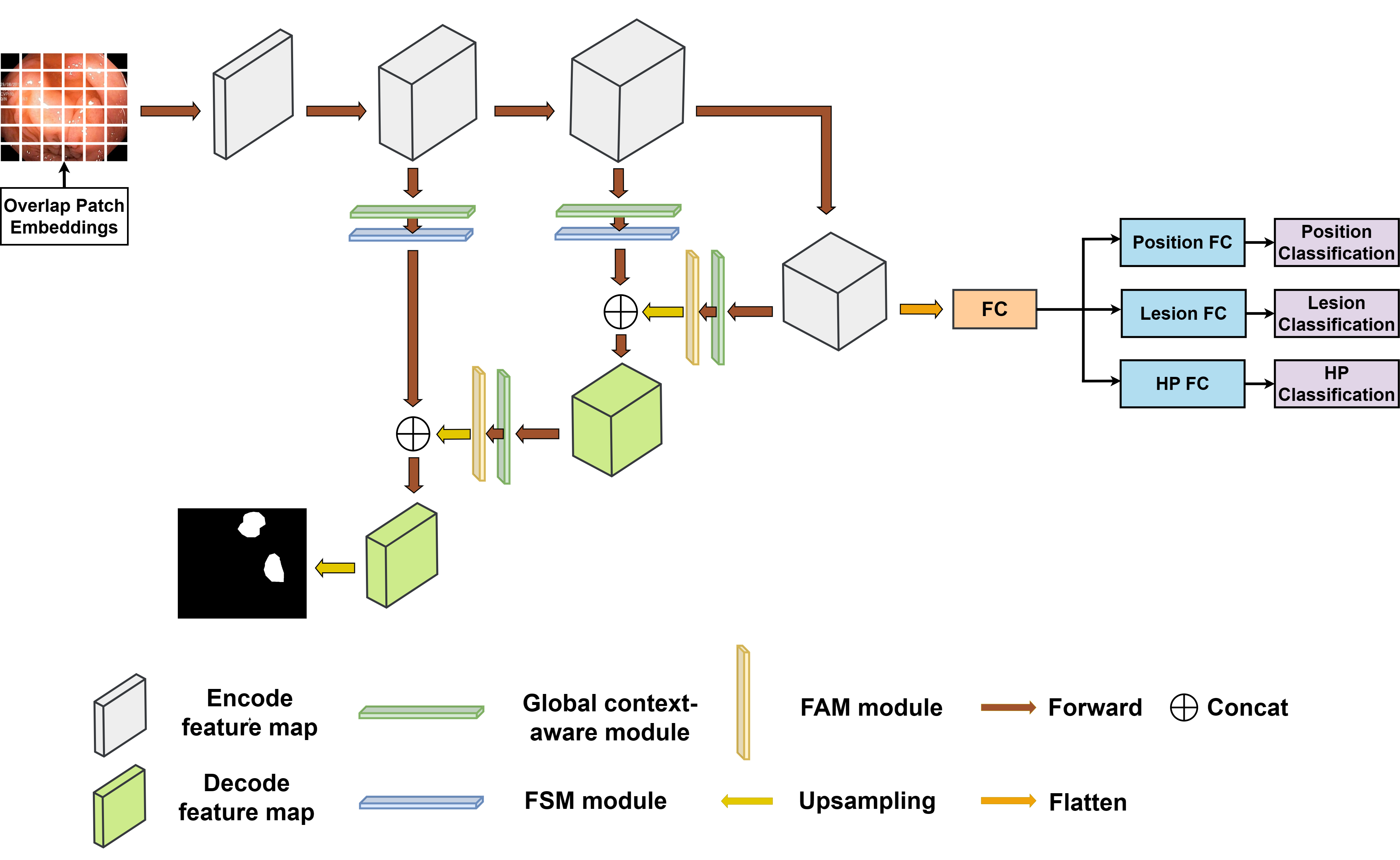}}
\caption{UGCANet architecture}
\label{fig:sfm-based-architecture}
\end{figure*}
\subsection{Overview}

The architecture of our proposed network, UGCANet, is illustrated in Fig. \ref{fig:sfm-based-architecture}. Our model uses the encoder-decoder design. 
The encoder is capable of acquiring a shared feature representation that is proficient in serving multiple tasks, producing four intermediate feature maps, and transmitting the final three to the CGNL \cite{cgnl} and module SE \cite{SE} to aggregate context information then a branch will undergo the average pooling layer and three FC layer to generate the classification labels for our tasks. Additionally, it will pass through the FaPN \cite{FaPN} module to generate the prediction output for the segmentation task.

\subsection{Backbone}

MiT \cite{segformer} improves upon ViT \cite{vit} by introducing hierarchical feature representation, overlapped patch merging, efficient self-attention, and Mix-FFN. The hierarchical feature representation generates multi-level features with different resolutions. Each hierarchical feature map $F{_i}$ has a resolution of ${\frac{H}{2^{i+1}} \times \frac{W}{2^{i+1}} \times C_i}$, where $i$ belongs to the set \{1, 2, 3, 4\} and $C{_i}$ increases with each level. While overlapped patch merging preserves local continuity in image patches. Efficient self-attention reduces computational load by using a reduction ratio. Mix-FFN mitigates the impact of zero padding on location information leakage, improving accuracy.

\subsection{Global Context-Aware Modules}

\subsubsection{Compact Generalized Non-Local}

The Compact Generalized Non-Local (CGNL) \cite{cgnl} module is a computer vision technique that models how different positions of an image interact across channels. Unlike the Non-Local module \cite{nonlocal}, CGNL applies distinct weights to each channel by first flattening the feature outputs after linear transformation layers. By using separate weights for each channel, it can better capture the underlying relationships between different parts of an image, leading to improved accuracy. CGNL stands out from channel attention modules due to the computational load required. Through the application of Taylor expansion, Yue et al. \cite{cgnl} are able to approximate the complex calculations required by the module with a simpler polynomial function. This makes the computation more efficient and helps to reduce the overall computational cost of the CGNL module.

\begin{figure*}[ht!]
\centering
\includegraphics[width=\textwidth]{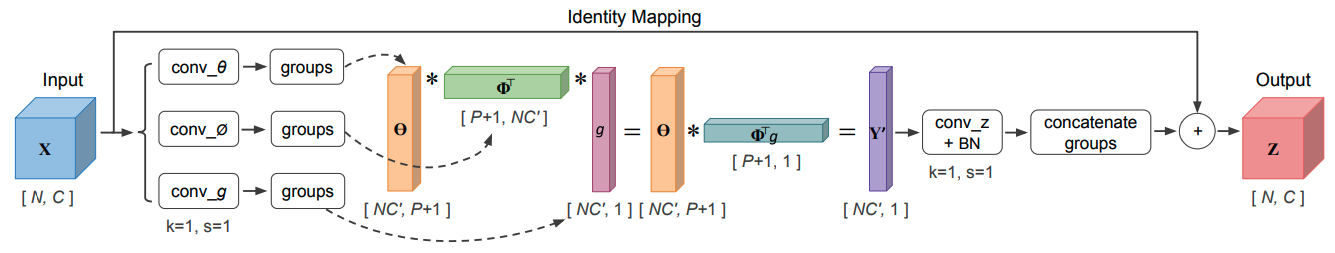} 
\caption{Grouped compact generalized non-local (CGNL) module \cite{cgnl}.}
\label{fig:CGNL}
\end{figure*}

\subsubsection{Squeeze and Excitation}

Squeeze and Excitation (SE) \cite{SE} is a neural network building block that improves the incantational power of CNNs by explicitly modeling interdependencies between channels of feature maps. 
The SE block consists of two main operations: the squeeze and excitation operations.

\begin{figure*}[ht!]
\centering
\includegraphics[width=\textwidth]{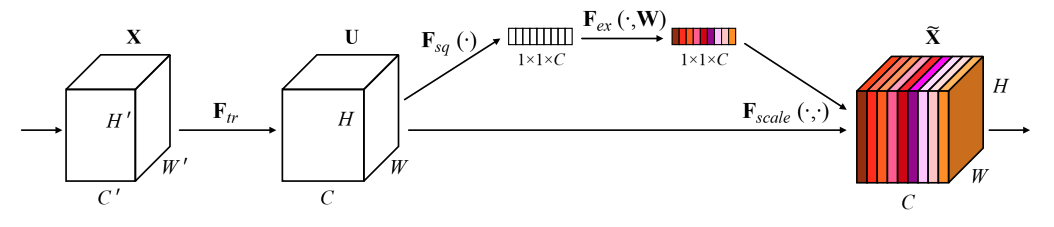} 
\caption{A Squeeze-and-Excitation block \cite{SE}.}
\label{fig:SE}
\end{figure*}

In the squeeze operation, a global spatial pooling operation is applied to the feature maps, which aggregates the information across spatial locations.

In the excitation operation, the channel-wise statistics obtained from the squeeze operation are used to compute a set of learnable weights. These weights are then used to modulate the feature maps through a gating mechanism, which assigns higher weights to informative channels and lower weights to less informative ones. The gating mechanism is typically implemented as a sigmoid or ReLU activation function.

Because of its simple and efficient design, it can be stacked multiple times to form a deep network.

\subsection{Decoder}

In the encoder-decoder architecture, the problem of object information loss can be particularly challenging for small objects in the image. Because the encoder compresses the image features into a lower-dimensional space, which can result in the loss of fine-grained details and spatial information. This can lead to reduced segmentation performance, especially in scenarios where small objects are of great importance, such as medical image analysis.

Huang et al. introduced the Feature-aligned Pyramid Network (FaPN) \cite{FaPN}, which is a new feature-aligned pyramid network for dense image prediction. The FaPN model addresses the issue of feature misalignment in the encoder-decoder architecture by using a feature alignment module that aligns the features extracted at different levels of the network.

\begin{figure*}[ht!]
\centerline{\includegraphics[width=0.7\textwidth]{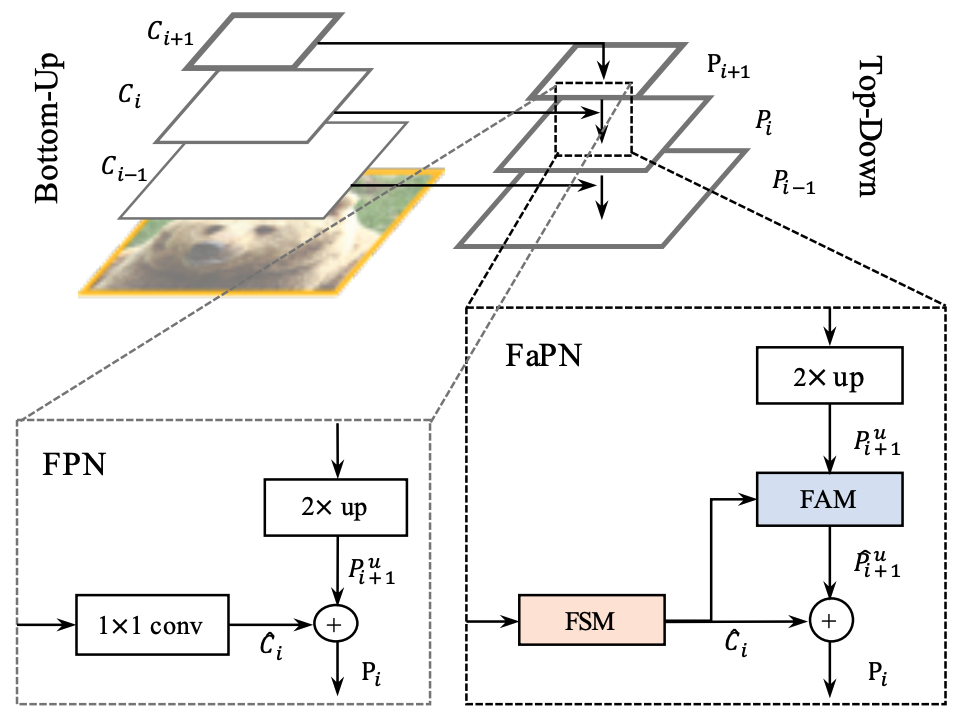}}
\caption{Overview comparison between FPN and FaPN \cite{FaPN}.}
\label{fig:fpn-vs-fapn}
\end{figure*}

\textbf{Feature Alignment Module}: The downsampling operations in the encoder-decoder architecture result in a misalignment between the feature maps of the encoder and decoder. This misalignment can cause issues when fusing the features through addition or concatenation near object boundaries. 

\begin{figure*}[ht!]
\centerline{\includegraphics[width=0.7\textwidth]{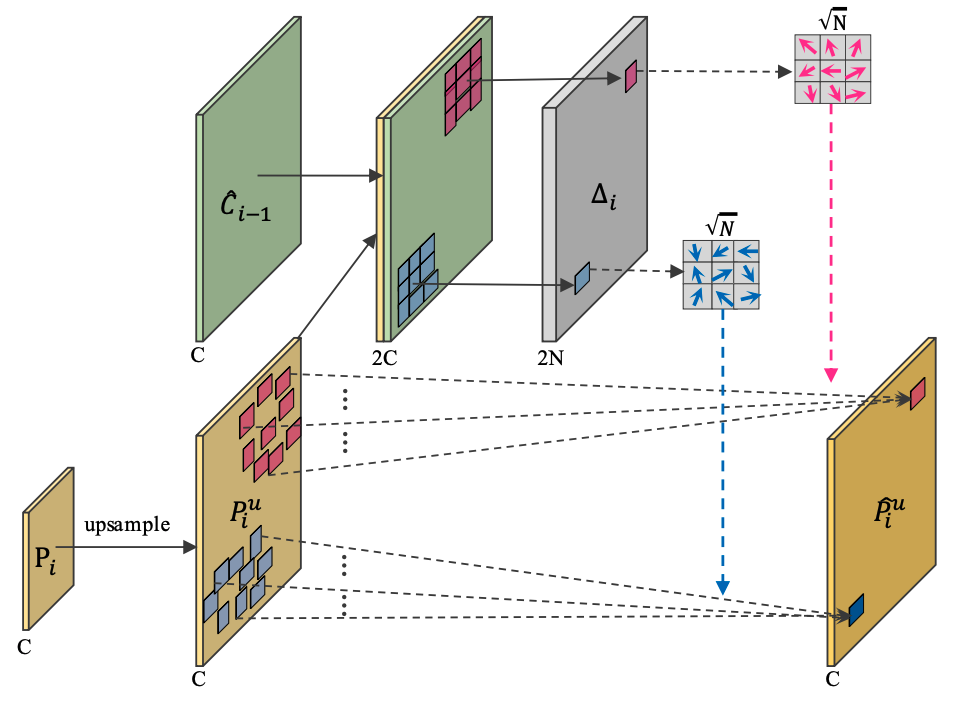}}
\caption{Feature alignment module \cite{FaPN}}
\label{fig:fam}
\end{figure*}

To address this, Feature Alignment Module (FAM) used deformable convolution and learnable offset fields to adjust the convolutional sample locations, which helps preserve accurate object boundaries. 

\textbf{Feature Selection Module}: Rather than employing a 1 × 1 convolution to reduce the channel dimension of intricate features, FSM (Feature Selection Module) adopts a weight assignment approach for feature maps. The FSM architecture is motivated by the SE module but differs in that it includes an additional shortcut between the input and scaled feature maps. This unique shortcut enables FSM to effectively enhance multi-scale feature aggregation. 


\subsection{Loss}

In our study, we evaluate two different types of datasets, including Upper GI and polyp data. In each task, we define a different type of loss to match the context.

\subsubsection{Loss for Colonoscopy}

Our loss function for this task is defined as: 
\begin{equation}
\label{eq:seg_loss}
\mathcal{L} = 
\mathcal{L}^{w}_{BCE} + 
\mathcal{L}^{w}_{IoU}
\end{equation}

where $\mathcal{L}^{w}_{BCE}$ is weighted binary cross-entropy and $\mathcal{L}^{w}_{IoU}$ is weighted IoU loss. Thus, $\mathcal{L}$ is called weighted BCE and IoU loss. In contrast to the standard binary cross-entropy loss function, the $\mathcal{L}^{w}_{BCE}$ assigns higher weights to hard pixels rather than treating all pixels equally. As for $\mathcal{L}^{w}_{IoU}$, increases the weights of hard pixels to emphasize their importance.

\subsubsection{Loss for upper GI tract}

We collect data from various sources to train our models, which we combine to create a comprehensive training dataset. However, in this merged dataset, each sample is only relevant to a subset of the tasks. To clarify the type of sample, we use $\mu_i^t \in \{0, 1\}$ as the indicator, with $t \in \{pos, le, hp, seg\}$ representing the tasks of anatomical site classification, lesion type classification, HP classification, and lesion segmentation, respectively. Suppose that $\mathbf{y}_i^t$ is the one-hot encoding of the label of the $i$-th sample in the $t$-th task. Assuming $\hat{\mathbf{y}}_i^t$ be the probabilistic output of the $i$-th sample in the $t$-th task. If $t \in \{pos, le, hp\}$ then $\mathbf{y}_i^t$ and $\hat{\mathbf{y}}_i^t$ are vectors whose length equals the number of classes, let $C_{pos}$ represent the number of anatomical sites, $C_{le}$ represent the number of lesion classes (including five lesion types and a negative one), and $C_{hp}$ represent whether the HP sample is positive or not. Specifically, $C_{pos} = 10$, $C_{le} = 6$, and $C_{hp} = 1$.
If $t=seg$, then both $\mathbf{y}_i^t$ and $\hat{\mathbf{y}}_i^t$ are two-dimensional matrices with dimensions equal to those of the input images.

The loss function $\mathcal{L}_{pos}$, used for the task of classifying anatomical sites, is a type of multi-class cross-entropy loss and is defined as follows:

\begin{equation}
\label{eq:lc1}
\mathcal{L}_{pos} = - \sum_{i=1}^{N} \left( \mu_i^{pos} \sum_{j=1}^{C_{pos}} y_i^{pos}(j) * log \: \hat{y}_i^{pos}(j)  \right)
\end{equation}
where $N$ is the number of training samples.

The loss $\mathcal{L}_{le}$ for lesion type classification task is another multi-class cross-entropy loss defined as follows:

\begin{equation}
\label{eq:lc2}
\mathcal{L}_{le} = - \sum_{i=1}^{N} \left( \mu_i^{le} \sum_{j=1}^{C_{le}} y_i^{le}(j) * log \: \hat{y}_i^{le}(j)  \right)
\end{equation}

The loss $\mathcal{L}_{hp}$ for HP classification is the binary cross-entropy loss defined as follows:

\begin{equation}
\label{eq:lc3}
\mathcal{L}_{hp} = - \sum_{i=1}^{N}  \mu_i^{hp} * \left(y_i^{hp} * log \: \hat{y}_i^{hp} + (1 - y_i^{hp}) * log \: (1 - \hat{y}_i^{hp})  \right)
\end{equation}

The total loss is a weighted combination of the above loss functions. This is defined as follows:
\begin{equation}
\label{eq:total_loss}
\mathcal{L}_{total} = \lambda_1 * \mathcal{L}_{pos} + \lambda_2 * \mathcal{L}_{le} + \lambda_3 * \mathcal{L}_{hp} + \lambda_4 * \mathcal{L}_{seg}
\end{equation}
where $\lambda_t$ indicates the importance level of the $t$-th task. In our work, we set $\lambda_1 = \lambda_2 = \lambda_3 = \lambda_4 = 1$.

\section{Experiments} \label{sec:Experiments}
\subsection{Datasets}
Our model is evaluated on two gastrointestinal datasets: one consists of images of the upper gastrointestinal tract, and the other is polyp data.

\subsubsection{Colonoscopy dataset}

For the task of polyp segmentation, we conducted our evaluation using a total of 5 datasets. Below, we provide detailed information about each of these datasets.

\textbf{Kvasir dataset} \cite{Kvasir}: This dataset comprises 1000 images with varying resolutions ranging from 720 × 576 to 1920 × 1072 pixels. Kvasir data is collected using endoscopic equipment at Vestre Viken Health Trust (VV) in Norway, annotated and verified by medical doctors (experienced endoscopists).

\textbf{CVC-ClinicDB dataset} \cite{Clinic}: CVC-ClinicDB is an openly available collection of 612 images extracted from 31 colonoscopy sequences, with a resolution of 384×288. The dataset is commonly utilized in medical image analysis, specifically in detecting polyps in colonoscopy videos through segmentation techniques.

\textbf{CVC-ColonDB dataset} \cite{Colon}: is provided by the Machine Vision Group (MVG). These images were extracted from 15 brief colonoscopy videos and consist of 380 images of 574 × 500 pixels resolution.

\textbf{CVC-T dataset} \cite{Endo}:  CVC-T dataset is a subset of a larger dataset named Endoscene, and it is primarily a test set. It is composed of 60 images that were obtained from 44 video sequences captured from 36 patients.

\textbf{ETIS-Larib dataset} \cite{Etis}: The dataset consists of 196 images with high resolution (1226 x 996).

\subsubsection{Upper GI tract dataset}
There are three datasets we collected from endoscopy findings of patients at the Institute of Gastroenterology and Hepatology and Hanoi Medical University Hospital: the first for anatomical site classification, the second for lesion segmentation and classification, and the last for HP classification. They are combined into a huge dataset of the upper GI tract.

\textbf{Anatomical site dataset}: This dataset includes 5546 images of 10 anatomical sites, all of which are captured directly from the endoscopic machine, including four lighting modes: WLI (White Light Imaging), FICE (Flexible spectral Imaging Color Enhancement), BLI (Blue Light Imaging), and LCI (Linked Color Imaging). The images in this dataset do not contain any lesions and have labels specifying the anatomical site. Table~\ref{table:site dataset} describes the details of this dataset.

\begin{table}[h!]
\centering
\caption{Number of images in each anatomical site and lighting mode}
\begin{tabular}{l|c c c c c} 
\hline
Anatomical site & WLI & FICE & BLI & LCI & TOTAL\\ 
\hline
\hline
Pharynx  & 177 & 134 & 120 & 119 & 550 \\ 
Esophagus & 169 & 141 & 116 & 127 & 553 \\
Cardia & 163 & 120 & 132 & 140 & 555\\
Gastric body & 174 & 135 & 124 & 120 & 553\\
Gastric fundus & 170 & 130 & 126 & 128 & 554\\
Gastric antrum & 155 & 143 & 131 & 125 & 554\\  
Greater curvature & 171 & 131 & 126 & 125 & 553\\  
Lesser curvature & 155 & 140 & 134 & 126 & 555\\  
Duodenum bulb & 156 & 141 & 135 & 128 & 560\\  
Duodenum & 163 & 138 & 127 & 131 & 559\\  
\hline
\end{tabular}
\label{table:site dataset}
\end{table}

\begin{figure*}[ht!]
\centerline{\includegraphics[width=\textwidth]{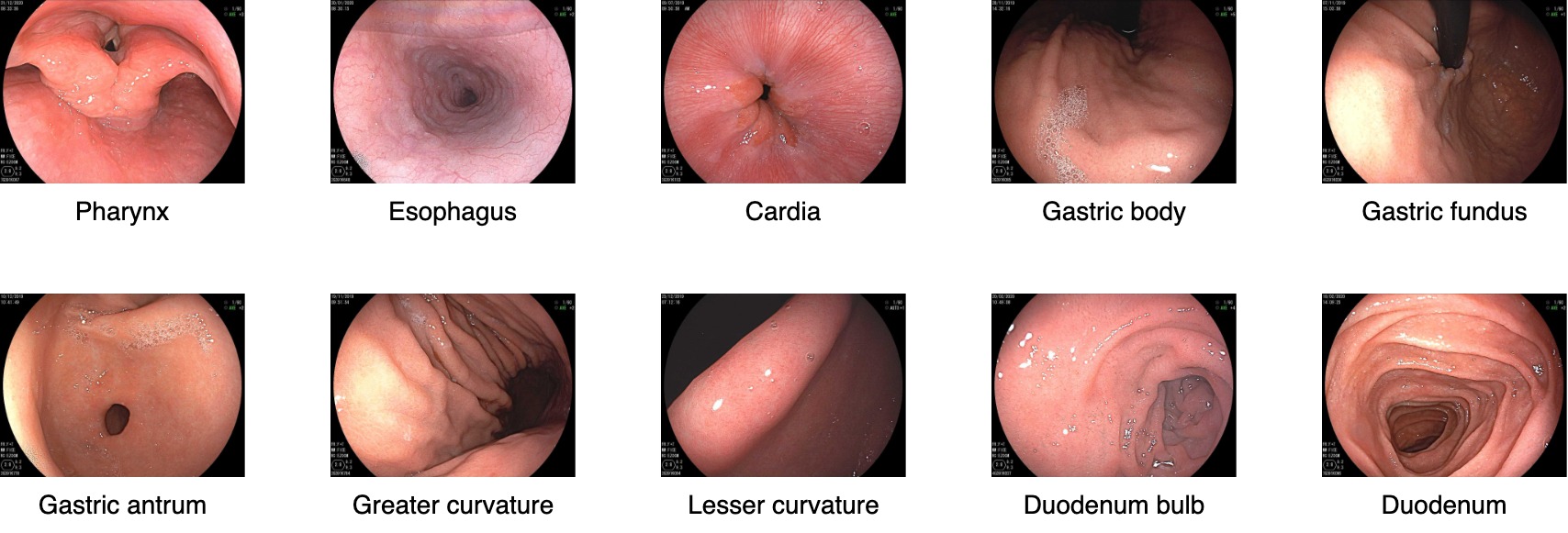}}
\caption{Some samples in anatomical dataset}
\label{fig:anatomical-dataset}
\end{figure*}

\textbf{Lesion dataset}: in this dataset, we have 4104 images of 5 types of lesions: reflux esophagitis, esophageal cancer, gastritis, stomach cancer, and duodenal ulcer. The images in this dataset have the annotations for both the classification and segmentation tasks. The numbers of images for reflux esophagitis, esophageal cancer, stomach cancer, and duodenal ulcer classes are 1335, 538, 1443, 538, and 250, respectively.

\begin{figure*}[ht!]
\centerline{\includegraphics[width=\textwidth]{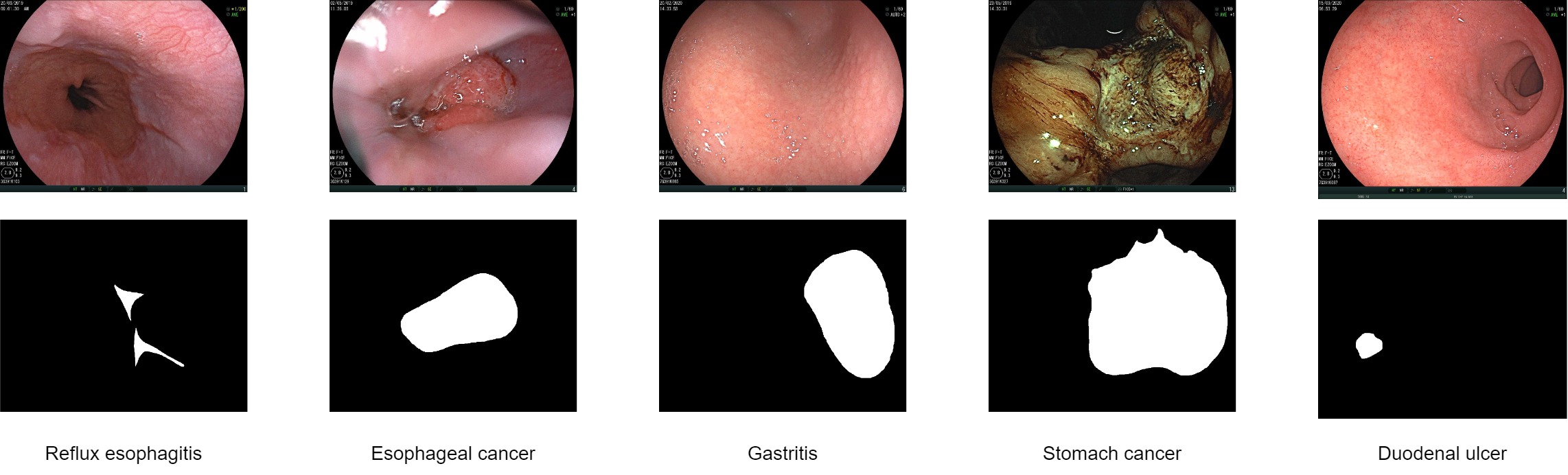}}
\caption{Some samples in lesion dataset}
\label{fig:lesion-dataset}
\end{figure*}

Figure \ref{fig:lesion-dataset} shows some samples in the lesion dataset.

\begin{figure*}[ht!]
\centerline{\includegraphics[width=\textwidth]{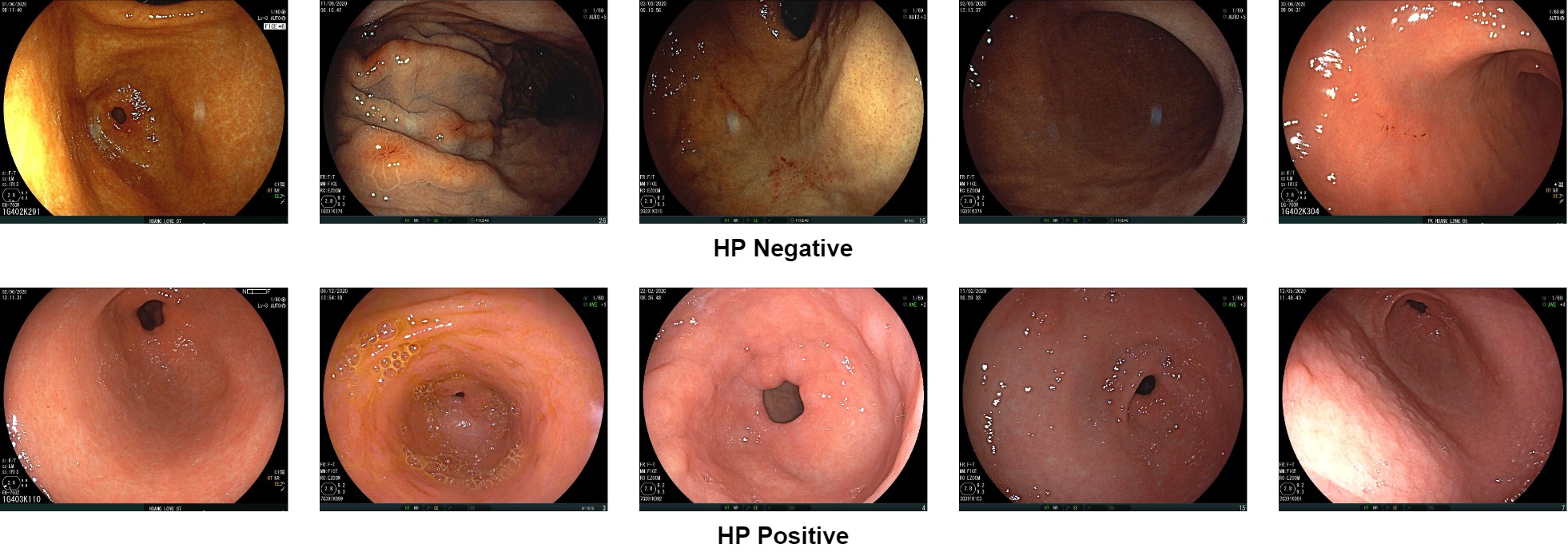}}
\caption{Some samples in HP dataset}
\label{fig:hp-dataset}
\end{figure*}

\textbf{HP dataset}: we have 1819 images in this dataset, including HP-positive and HP-negative images. Figure \ref{fig:hp-dataset} are some samples in the HP dataset.

\subsection{Implementation Details}

We performed two separate experiments on the upper GI and polyp datasets to validate our proposed UGCANet. 

\begin{itemize}
\item\textbf{For polyp dataset}: For the polyp dataset, our approach utilizes an image size of 384x384 and employs multi-scale training with size ratios of {0.75, 1, 1.25}, respectively. Regarding the optimization algorithm, we adopt Adam with an initial learning rate of 1e-4. Our data augmentation strategy comprises  Flip, HueSaturation, and RandomBrightnessContrast, each with a probability of 0.5. 
We perform three experiments with different dataset setups to compare with SoTA models.

\textbf{Experiment 1} \label{polyp:exp1}: The splitting method recommended in \cite{pranet} is applied, where 90\% of the Kvasir and ClinicDB datasets are allocated for training. The remaining images in Kvasir and CVC-ClinicDB datasets, along with all images from CVC-ColonDB, CVC-T, and ETIS-Larib, are used for testing.

\textbf{Experiment 2}: 5-fold cross-validation on the CVC-ClinicDB and Kvasir datasets.

\textbf{Experiment 3}: Cross-dataset evaluation with three training-testing configurations:
\begin{enumerate}
\item CVC-ColonDB and ETIS-Larib for training, CVC-ClinicDB for testing;
\item CVC-ColonDB for training, CVC-ClinicDB for testing;
\item CVC-ClinicDB for training, ETIS-Larib for testing.
\end{enumerate}

\item\textbf{For upper GI dataset}: 

5-fold cross-validation schema. The datasets are first divided into five subfolds, which are then merged to create larger folds. In the anatomical site dataset, each subfold consists of an equal number of images per anatomical site and lighting mode. Similarly, the lesion dataset's subfolds contain an equal number of images per lesion type, and the HP dataset's folds contain the same number of HP positive and HP negative samples. Additionally, a marker vector $\mu$ is generated to denote the sample type.

We use two variants of the MiT backbone, namely MiT-B2 and MiT-B3, to conduct training on images of size 480x480 without utilizing multi-scale techniques. The Adam optimizer is utilized with a learning rate linear warmup and cosine strategy annealing. 

The following experiments are performed to evaluate the model's performance:

    \textbf{Experiment 4}: we evaluate the impact of two MiT configurations on UGCANet, including MiT-B2 and MiT-B3, and compare their performance with backbones of EndoUNet \cite{manh2022endounet}, including VGG19, ResNet50, and DenseNet121.
    
    \textbf{Experiment 5}: in the classification tasks, we train the single-tasking instances of models, including VGG19, ResNet50, DenseNet121, and MiT-B3, each of them trained on separate data. We compare the performance of multi-tasking models and the single-tasking models.
    
    \textbf{Experiment 6}: in the lesion segmentation task, we train five single-tasking instances of EndoUNet and five single-tasking instances of UGCANet, each of which trained on separate lesion data. Then, we compare the performance of multi-tasking models versus the single-tasking instances.
\end{itemize}

All training is done on a machine with a 3.7GHz AMD Ryzen 3970X CPU, 128GB RAM, and an NVIDIA GeForce GTX 3090 GPU.

\subsection{Results and Discussion}

\subsubsection{Comparison with SoTA methods}

\begin{itemize}
\item\textbf{For polyp dataset}:

\begin{table}[!ht]
\caption{Performance comparison of different methods on the Kvasir, ClinicDB, ColonDB, CVC-T and ETIS-Larib test sets. All results of UGCANet are averaged over five runs.}
\centering
{\renewcommand{\arraystretch}{1.2}
\resizebox{1\textwidth}{!}{%
\begin{tabular}{c|cc|cc|cc|cc|cc}
\hline
Method & \multicolumn{2}{c|}{Kvasir} & \multicolumn{2}{c|}{CVC-ClinicDB} & \multicolumn{2}{c|}{CVC-ColonDB} & \multicolumn{2}{c|}{CVC-T} & \multicolumn{2}{c}{ETIS-Larib}  \\
\cline{2-11}
& mDice & mIOU               & mDice & mIOU                 & mDice & mIOU                & mDice & mIOU                   & mDice & mIOU              \\
\hline
\hline
UNet \cite{Unet}           & 0.818 & 0.746 & 0.823 & 0.750 & 0.512 & 0.444 & 0.710 & 0.627  & 0.398 & 0.335 \\
UNet++ \cite{Unet++}         & 0.821 & 0.743 & 0.794 & 0.729 & 0.483 & 0.410 & 0.707 & 0.624  & 0.401 & 0.344 \\
PraNet \cite{pranet}         & 0.898 & 0.840 & 0.899 & 0.849 & 0.709 & 0.640 & 0.871 & 0.797  & 0.628 & 0.567 \\
HarDNet-MSEG \cite{hardnet_mseg}  & 0.912 & 0.857 & 0.932 & 0.882 & 0.731 & 0.660 & 0.887 & 0.821  & 0.677 & 0.613 \\
CaraNet \cite{caranet}          & 0.918 & 0.865 & 0.936 & 0.887 & 0.773 & 0.689 & 0.903 & 0.838& 0.747 & 0.672\\
TransUNet \cite{transunet}   & 0.913 & 0.857 & 0.935 & 0.887 & 0.781 & 0.699 & 0.893 & 0.824 & 0.731 & 0.660   \\
TransFuse-L* \cite{transfuse}   & 0.920 & 0.870 &0.942 & \textbf{0.897} & 0.781 & 0.706 & 0.894 & 0.826 & 0.737 & 0.663\\

ColonFormer-S \cite{ColonFormer} & 0.927 & 0.877 & 0.932 & 0.833 & 0.811 & 0.730 & 0.894 & 0.826 & 0.789 & 0.711\\

ColonFormer-L \cite{ColonFormer}   & 0.924 &0.876 & 0.932 & 0.884 & 0.811 & 0.733 & 0.906 & 0.842 & 0.801 & 0.722\\


\textit{\textbf{UGCANet (Ours)}} & \textbf{0.928} & \textbf{0.881} & \textbf{0.943} & \underline{0.896} & 
\textbf{0.827} & \textbf{0.749} & \textbf{0.910} & \textbf{0.847} & \textbf{0.822} & \textbf{0.744} \\[2pt]
\hline
\end{tabular}%
}
}
\label{tab:sota}
\end{table}

Table \ref{tab:sota} shows the comparison result for Experiment 1. Our model UGCANet outperforms previous SoTA models in mDice and mIoU metrics. Despite the impressive performance of ColonFormer-S on the Kvasir dataset, our model UGCANet outperforms ColonFormer-S and ColonFormer-L on the ETIS-Larib dataset by 2.1\% in mDice and 2.2\% on mIoU. On CVC-ColonDB, UGCANet surpasses TransFuse-L* and TransUnet with 4.6\% mDice and also mIoU of 5\% and 4.3\%, respectively. However, on our CVC-ClinicDB dataset, our model only approximates mDice and mIoU compared to TransFuse-L*.

Table \ref{tab:kfold} shows us that UGCANet outperforms all state-of-the-art models in mDice, mIoU, precision, and even recall in the 5-fold cross-validation experiment on the ClinicDB dataset. With the Kvasir dataset, we continue to outperform other models on mDice and precision metrics. However, our model is 0.8\% worse in recall and 0.1\% in mIoU. 
It is worth mentioning that our UGCANet model has displayed remarkable stability on both datasets, with a relatively low standard deviation.

\begin{table}[!ht]
\caption{Performance comparison of different methods on 5-fold cross-validation of the CVC-ClinicDB and Kvasir datasets. All results are averaged over 5 folds.}
\centering
{\renewcommand{\arraystretch}{1.2}
\begin{tabular}{c|c|cccc}
\hline
Dataset & Method & mDice & mIOU  & Recall & Precision  \\
\hline
\hline
\multirow{9}{*}{\rotatebox[origin=c]{90}{ClinicDB}} 

& ResUNet++ \cite{resunet++}  & $0.815 \pm 0.018$ & $0.736 \pm 0.017$ & $0.832\pm 0.018$ &  $0.830 \pm 0.020$ \\
& DoubleUNet \cite{DoubleUnet} & $0.920 \pm 0.018$ & $0.866 \pm 0.025$ & $0.922 \pm 0.027$ & $0.928 \pm 0.017$ \\
& DDANet \cite{ddanet} & $0.860 \pm 0.014$ & $0.786 \pm 0.017$ & $0.858 \pm 0.023$ & $0.892 \pm 0.014$ \\
& ColonSegNet \cite{colonsegnet} & $0.817 \pm 0.020$ & $0.873 \pm 0.024$ & $0.926 \pm 0.025$ & $0.933 \pm 0.014$ \\
& HarDNet-MSEG \cite{hardnet_mseg} & $0.923 \pm 0.020$ & $0.873 \pm 0.024$ & $0.926 \pm 0.025$ & $0.933 \pm 0.014$ \\
& PraNet \cite{pranet} & $0.933 \pm 0.012$ & $0.884 \pm 0.015$ & $0.940 \pm 0.005$ & $0.937 \pm 0.016$ \\

& ColonFormer-S \cite{ColonFormer} & $0.948 \pm 0.002$ & $0.904 \pm 0.004$ & $0.958 \pm 0.003$ & $0.941 \pm 0.004$ \\

& ColonFormer-L \cite{ColonFormer} & $0.947 \pm 0.002$ & $0.903 \pm 0.003$ & $0.956 \pm 0.002$ & $0.942 \pm 0.005$ \\

& \textit{\textbf{UGCANet (Ours)}} & \textbf{0.950 $\pm$ 0.001} & \textbf{0.907 $\pm$ 0.003} & \textbf{0.959 $\pm$ 0.007} & \textbf{0.947 $\pm$ 0.012} \\
[2pt]
\hline
\hline
\multirow{8}{*}{\rotatebox[origin=c]{90}{Kvasir}} 

& ResUNet++ \cite{resunet++}            & $0.780\pm0.010$          & $0.681\pm0.008$         & $0.834\pm0.010$  & $0.799\pm0.010$      \\
& DoubleUNet \cite{DoubleUnet} & $0.879 \pm 0.018$ & $0.816 \pm 0.026$ & $0.902 \pm 0.027$ & $0.894 \pm 0.039$ \\
& DDANet \cite{ddanet} & $0.860 \pm 0.005$ & $0.791 \pm 0.004$ & $0.876 \pm 0.015$ & $0.892 \pm 0.018$ \\
& ColonSegNet \cite{colonsegnet} & $0.676 \pm 0.037$ & $0.557 \pm 0.040$ & $0.731 \pm 0.088$ & $0.730 \pm 0.080$ \\
& HarDNet-MSEG \cite{hardnet_mseg} & $0.889 \pm 0.011$ & $0.831 \pm 0.011$ & $0.892 \pm 0.015$ & $0.926 \pm 0.014$ \\
& PraNet \cite{pranet} & $0.883\pm0.020$          & $0.822\pm0.020$         & $0.897\pm0.020$  & $0.906\pm0.010$      \\

& ColonFormer-S \cite{ColonFormer} & 0.924 $\pm$ 0.008 & \textbf{0.875 $\pm$ 0.010} & \textbf{0.941 $\pm$ 0.010} & 0.927 $\pm$ 0.008 \\

& ColonFormer-L \cite{ColonFormer} & 0.917 $\pm$ 0.006 & 0.865 $\pm$ 0.007 & 0.932 $\pm$ 0.007 & 0.926 $\pm$ 0.008 \\

& \textit{\textbf{UGCANet (Ours)}} & \textbf{0.926 $\pm$ 0.002} & \underline{0.874 $\pm$ 0.003} & \underline{0.933 $\pm$ 0.004} & \textbf{0.934 $\pm$ 0.003} \\[2pt]
\hline
\end{tabular}%
}
\label{tab:kfold}
\end{table}

\begin{figure*}[!ht]
\centerline{\includegraphics[width=\textwidth]{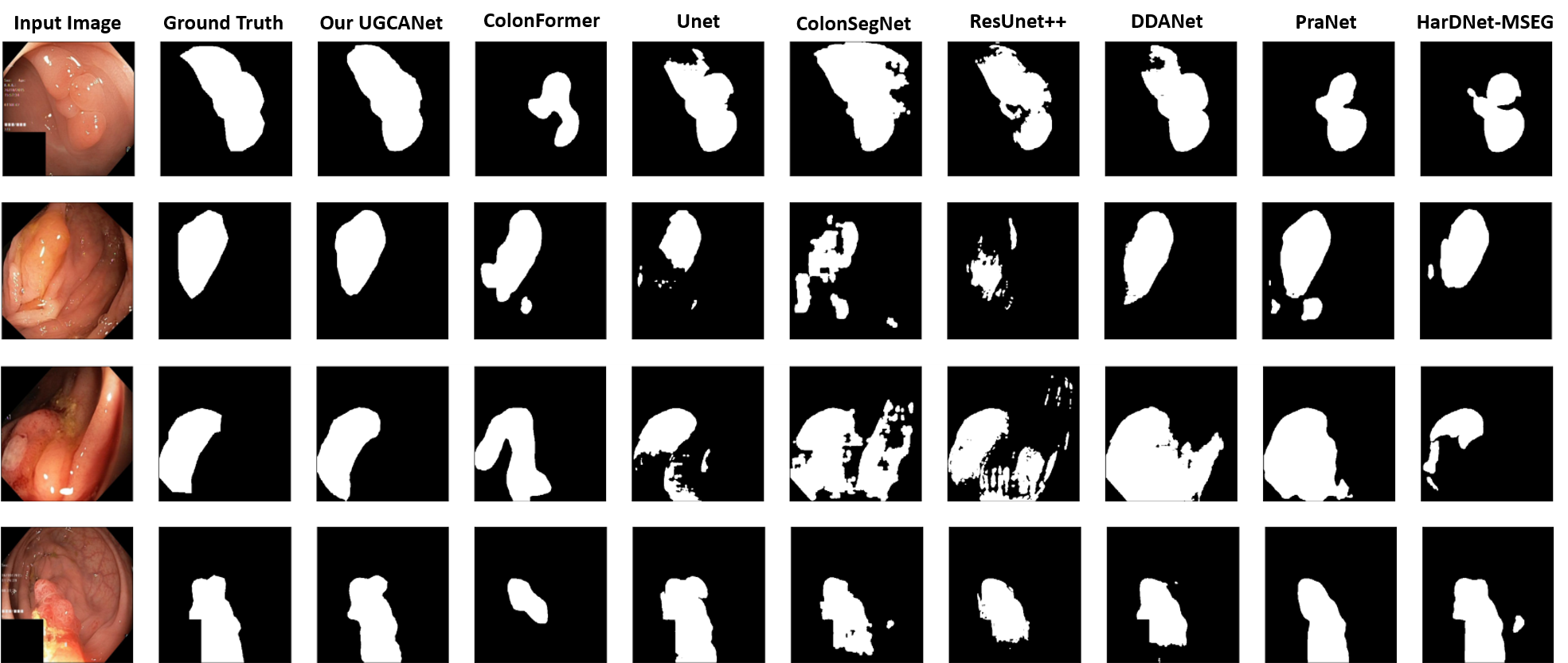}}
\caption{Qualitative result comparison of different models trained on the first fold of the 5-fold cross-validation on the
Kvasir dataset.}
\label{fig:Exp6}
\end{figure*}

\begin{figure*}[ht!]
\centerline{\includegraphics[width=\textwidth]{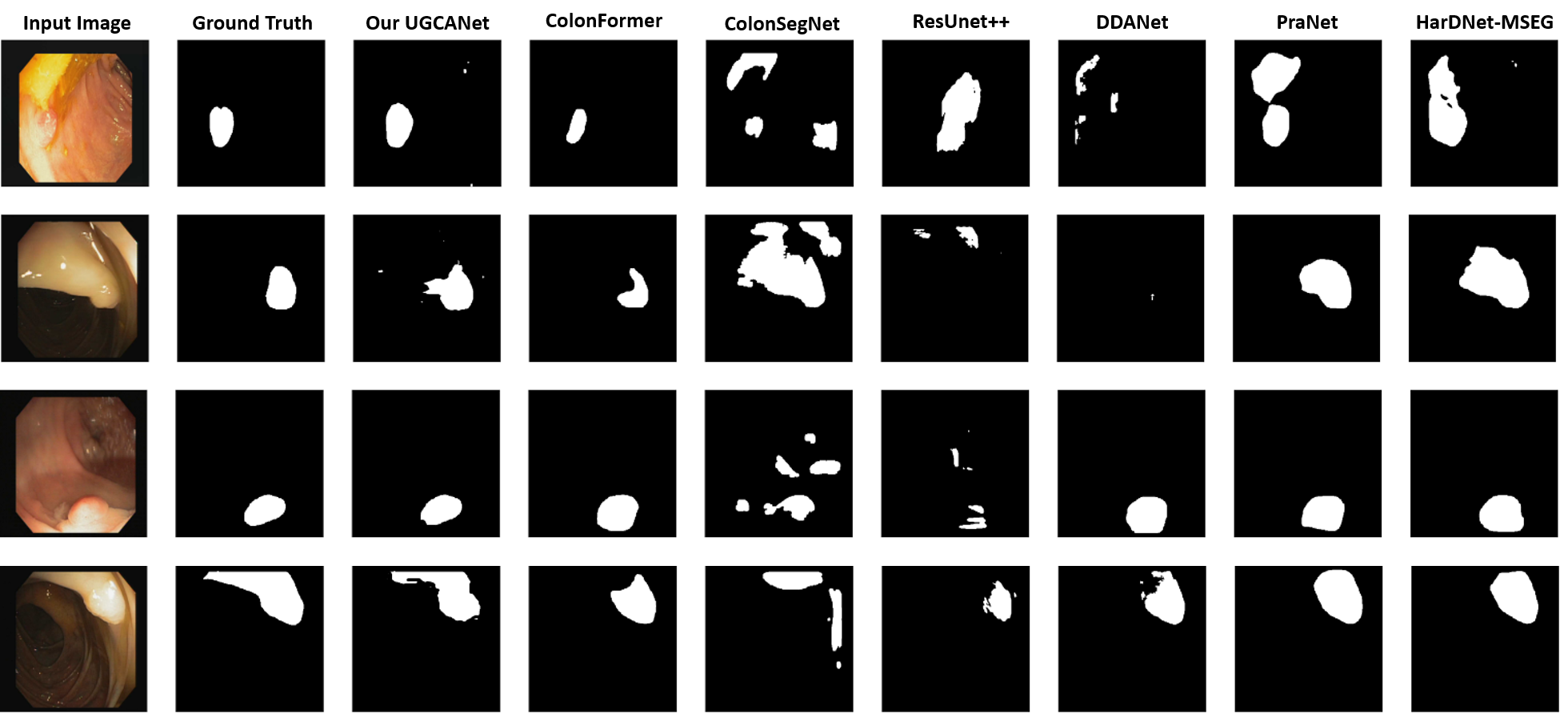}}
\caption{Qualitative result comparison using CVC-Colon for training and CVC-Clinic for testing.}
\label{fig:Exp7}
\end{figure*}

%

Table \ref{tab:complexity} compares UGCANet and other benchmark models regarding their size and computational complexity. Our model exhibits greater parameters than CNN-based models like CaraNet \cite{caranet}, or HarDNet-MSEG \cite{hardnet_mseg}. However, compared to the corresponding transformer architectures, our model proves advantageous in terms of parameter optimization. Despite having only a slightly higher computational complexity of 2.93 GFLOPs than ColonFormer-S \cite{ColonFormer}, our model still outperforms it in terms of overall model performance.

\begin{table*}
\caption{Performance comparison of different methods on cross-dataset configurations. All results are averaged over five runs.}
\centering

{\renewcommand{\arraystretch}{1.2}
\begin{tabular}[!ht]{cc|c|c|cccc}
\hline
\multicolumn{2}{c|}{Train} & Test & Method & mDice & mIOU  & Recall & Precision  \\
\hline
\hline
\multirow{7}{*}{\rotatebox[origin=c]{90}{CVC-ColonDB}} & \multirow{7}{*}{\rotatebox[origin=c]{90}{+ ETIS-Larib}}  &  \multirow{7}{*}{\rotatebox[origin=c]{90}{CVC-ClinicDB}} 
& ResUNet++ \cite{resunet++} & 0.406 & 0.302 & 0.481 & 0.496 \\
&& & ColonSegNet \cite{colonsegnet} & 0.427 & 0.321 & 0.529 & 0.552 \\
&& & DDANet \cite{ddanet} & 0.624 & 0.515 & 0.697 & 0.692 \\
&& & DoubleUNet \cite{DoubleUnet} & 0.738 &  0.651 & 0.758 & 0.824 \\
&& & HarDNet-MSEG \cite{hardnet_mseg} & 0.765 &  0.681 & 0.774 & 0.863 \\
&& & PraNet \cite{pranet} & 0.779 &  0.689 & 0.832 & 0.812 \\
&& & ColonFormer-S \cite{ColonFormer} & \textbf{0.851} &  0.771 & 0.853 & 0.896 \\

&& & ColonFormer-L \cite{ColonFormer} & 0.847 &  0.770 & 0.844 & \textbf{0.902} \\

&& & \textit{\textbf{UGCANet (Ours)}}     &   \underline{0.849}  &   \textbf{0.773}  &   \textbf{0.855}   &  \underline{0.893} \\[2pt]
\hline
\hline
\multirow{9}{*}{\rotatebox[origin=c]{90}{CVC-ColonDB}} & &  \multirow{9}{*}{\rotatebox[origin=c]{90}{CVC-ClinicDB}} 
& ResUNet++ \cite{resunet++} & 0.339 & 0.247 & 0.380 & 0.484 \\
&& & DoubleUNet \cite{DoubleUnet} & 0.441 &  0.375 & 0.423 & 0.639 \\
&& & DDANet \cite{ddanet} & 0.476 & 0.370 & 0.501 & 0.644 \\

&& & ColonSegNet \cite{colonsegnet} & 0.582 &    0.268 & 0.511 & 0.460 \\
&& & HarDNet-MSEG \cite{hardnet_mseg} & 0.721 &  0.633 & 0.744 & 0.818 \\
&& & PraNet \cite{pranet} & 0.738 & 0.647 & 0.751 & 0.832 \\

&& & ColonFormer-S \cite{ColonFormer} & \textbf{0.816} &  \textbf{0.731} & 0.809 & \textbf{0.881} \\

&& & ColonFormer-L \cite{ColonFormer} & 0.804 &  0.723 & 0.794 & 0.877 \\

&& & \textit{\textbf{UGCANet (Ours)}} & \underline{0.809}   &  \underline{0.724} & \textbf{0.822}  & \underline{0.870}\\[2pt]
\hline
\hline
\multirow{10}{*}{\rotatebox[origin=c]{90}{CVC-ClinicDB}} & &  \multirow{10}{*}{\rotatebox[origin=c]{90}{ETIS-Larib}} 
& ResUNet++ \cite{resunet++} & 0.211 & 0.155 & 0.309 & 0.203 \\
&& & ColonSegNet \cite{colonsegnet} & 0.217 &    0.110 & 0.654 & 0.144 \\
&& & DDANet \cite{ddanet} & 0.400 & 0.313 & 0.507 & 0.464 \\
&& & DoubleUNet \cite{DoubleUnet} & 0.588   & 0.500         & 0.689  & 0.599      \\
&& & PraNet \cite{pranet} & 0.631 &    0.555 &  0.762  &    0.597\\
&& & HarDNet-MSEG \cite{hardnet_mseg} & 0.659 &  0.583 & 0.676 & 0.705 \\
&& & ColonFormer-S \cite{ColonFormer} & 0.723 &  0.635 & 0.797 & 0.731 \\

&& & ColonFormer-L \cite{ColonFormer} & 0.760 &  0.673 & 0.859 & 0.734 \\

&& & \textit{\textbf{UGCANet (Ours)}}  & \textbf{0.803}    &   \textbf{0.722} & \textbf{0.905}  & \textbf{0.766}\\[2pt]
\hline
\end{tabular}
}
\label{tab:cross-dataset}
\end{table*}

\begin{table*}[!ht]
\caption{Number of parameters and GFLOPs of different methods}
\centering
{\renewcommand{\arraystretch}{1.2}
\begin{tabular}{c|cc}
\hline
Method & Parameters (M) & GFLOPs\\
\hline
\hline
PraNet \cite{pranet}  & 32.55 &   13.11        \\
HarDNet-MSEG \cite{hardnet_mseg} & 33.34 & 11.38         \\
CaraNet \cite{caranet} & 46.64 & 21.69       \\
TransUNet \cite{transunet} & 105.5 & 60.75         \\
TransFuse-L* \cite{transfuse} & - & -          \\
SegFormer-B3 \cite{segformer} & 47.22 & 33.68 \\
MiT-B3-FaPN & 46.54 & 16.85 \\
ColonFormer-S \cite{ColonFormer} & 33.04 & 16.03 \\
ColonFormer-L \cite{ColonFormer} & 52.94 & 22.94 \\
\textit{\textbf{UGCANet-S~(Ours)}} & 46.22 & 18.96 \\

\hline
\end{tabular}
}
\label{tab:complexity}
\end{table*}

\item\textbf{For upper GI dataset}:
\end{itemize}

Our study investigated the effectiveness of various backbones in classifying lesions. According to Table \ref{table:acc}, we found that model performance across datasets exhibits minimal variability. Notably, two datasets, anatomical site and lesion, demonstrated high accuracy rates. Even the lowest accuracies achieved by the single-tasking VGG19 model for anatomical site classification, lesion classification, and HP classification tasks were impressive at 97.07\%, 98.51\%, and 91.21\%, respectively. However, our results indicate that multi-tasking models outperform single-tasking models. Particularly, SFMNet with MiT-B3 as the backbone and EndoUNet with Resnet50 as the backbone proved to be the top-performing models across all three tasks, achieving impressive accuracy rates of 98.46\%, 99.63\%, and 93.46\%, respectively.

\begin{table}[!ht]
\centering
\def\arraystretch{1.2}
\caption{Accuracy comparison on the three classification tasks}
\resizebox{1.0\textwidth}{!}{%
\begin{tabular}{c | c c c c}
\hline
Method & Backbone & Anatomical site & Lesion classification & HP classification \\ 
 &  & classification &  &  \\ 
\hline
\hline
VGG19 (cls only)           &   VGG19            & $97.07 \pm 0.29 \%$ & $98.51 \pm 0.69 \%$ & $91.21 \pm 1,27 \%$ \\
Resnet50 (cls only)        &   Resnet50          & $97.53 \pm 0.29 \%$ & $98.79 \pm 1.09 \%$ & $91.87 \pm 1.08 \%$ \\
DenseNet121 (cls only)     &   DenseNet121          & $97.65 \pm 0.29 \%$ & $99.16 \pm 0.76 \%$ & $91.81 \pm 1.23 \%$ \\
MiT-B3 (cls only)          &  MiT-B3           & $97.58 \pm 0.86 \%$ & $99.45 \pm 0.36 \%$ & $91.43 \pm 1.15 \%$ \\
\hline
\hline
\multirow{3}{*}{EndoUNet}  & VGG19                      & $98.09 \pm 0.30 \%$ & $99.58 \pm 0.44 \%$             & $93.13 \pm 1.02 \%$           \\
& ResNet50                   & $98.00 \pm 0.49 \%$ & $\mathbf{99.63 \pm 0.26 \%}$    & $\mathbf{93.46 \pm 0.83 \%}$  \\ 
& DenseNet121                & $98.28 \pm 0.50 \%$ & $99.44 \pm 0.75 \%$             & $93.19 \pm 1.14 \%$           \\
\hline
\hline
\multirow{2}{*}{UGCANet} & MiT-B2                     & $98.30 \pm 0.31 \%$           & $99.11 \pm 0.76 \%$ & $93.35 \pm 0.79 \%$ \\
& MiT-B3                     & $\mathbf{98.46 \pm 0.41 \%}$  & $99.54 \pm 0.65 \%$ & $93.29 \pm 0.82 \%$ \\
\hline
\end{tabular}%
}
\label{table:acc}
\end{table}

Table \ref{table:dice} exhibits the segmentation task results for the upper digest dataset. The multi-tasking model demonstrated superior performance to its single-tasking counterpart in most tests. Our investigation indicated that multi-tasking learning was notably more effective, particularly when utilizing the UGCANet model. 

\begin{table}[!ht]
\def\arraystretch{1.2}
\centering
\caption{Dice Score comparison on the segmentation task}
\resizebox{1\textwidth}{!}{%

\begin{tabular}{c|c c c c c c}
\hline
Method                      & Backbone      & \multicolumn{1}{c}{\begin{tabular}[c]{@{}c@{}}Reflux\\ esophagitis\end{tabular}}   & \multicolumn{1}{c}{\begin{tabular}[c]{@{}c@{}}Esophageal\\ cancer\end{tabular}}   & \multicolumn{1}{c}{\begin{tabular}[c]{@{}c@{}}Duodenal\\ ulcer\end{tabular}}    & Gastritis        & \multicolumn{1}{c}{\begin{tabular}[c]{@{}c@{}}Stomach\\ cancer\end{tabular}} \\ 
\hline
\hline
UNet (seg only)             & ResNet50      & $0.457 \pm 0.011$          & $0.807 \pm 0.005$          & $0.709 \pm 0.021$          & $0.444 \pm 0.057$          & $0.854 \pm 0.021$             \\
SFM-based (seg only)        & MiT-B3        & $0.515 \pm 0.002$          & $0.839 \pm 0.012$          & $\mathbf{0.737 \pm 0.008}$ & $0.477 \pm 0.051$          & $\mathbf{0.896 \pm 0.009}$    \\
\hline
\hline
\multirow{3}{*}{EndoUNet}   & VGG19         & $0.462 \pm 0.014$          & $0.807 \pm 0.006$          & $0.648 \pm 0.024$          & $0.419 \pm 0.048$          & $0.851 \pm 0.009$             \\
& ResNet50      & $0.464 \pm 0.006$          & $0.819 \pm 0.009$          & $0.676 \pm 0.024$          & $0.443 \pm 0.065$          & $0.860 \pm 0.009$             \\
& DenseNet121   & $0.474 \pm 0.008$          & $0.824 \pm 0.007$          & $0.670 \pm 0.014$          & $0.457 \pm 0.066$          & $0.866 \pm 0.014$             \\
\hline
\hline
\multirow{2}{*}{UGCANet}  & MiT-B2        & $0.493 \pm 0.021$         & $0.837 \pm 0.012$         & $0.704 \pm 0.025$          & $0.476 \pm 0.074$          & $0.885 \pm 0.007$             \\
& MiT-B3        & $\mathbf{0.517 \pm 0.007}$ & $\mathbf{0.847 \pm 0.012}$ & $0.723 \pm 0.009$          & $\mathbf{0.502 \pm 0.072}$ & $0.892 \pm 0.008$             \\
\hline
\end{tabular}%
}

\label{table:dice}
\end{table}

\begin{figure*}[!ht]
\centerline{\includegraphics[width=\textwidth]{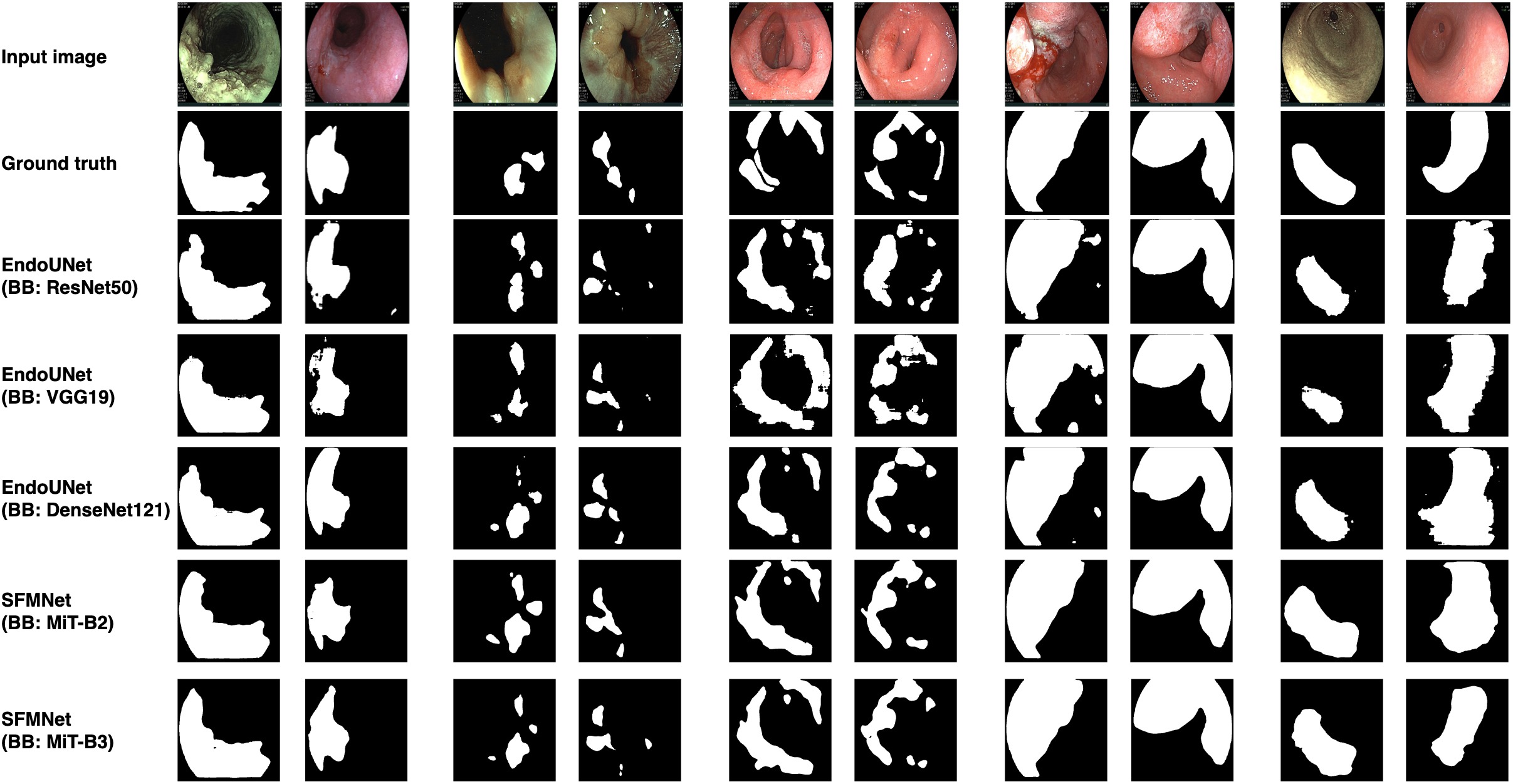}}
\caption{Some examples of the lesion segmentation task}
\label{fig:lesion-segmentation}
\end{figure*}

\subsubsection{Ablation study}

\textbf{Effectiveness of the FaPN}: At first glance, it is apparent that both SegFormer-B3 and MiT-B3-FaPN utilize the same backbone, MiT-B3, differing only in their respective decoders. Notably, the average mDice score of MiT-B3-FaPN is higher than that of SegFormer-B3, and the parameters and GFLOPs values displayed in Table \ref{tab:complexity} are significantly lower than those of MiT-B3-FaPN. The comparison results show that the computational cost with FaPN is much lower than that of SegFormer-B3 and the results are also slightly improved, so we choose FaPN as the decoder.

\begin{table*}[!ht]

\caption{Ablation study on the effectiveness of the modules with \textbf{Experiment 1 of polyp dataset} setup. All results are averaged over five runs.}
\centering
{\renewcommand{\arraystretch}{1.2}
\resizebox{1\textwidth}{!}{%
\begin{tabular}{c|cc|cc|cc|cc|cc}
\hline
Method  & \multicolumn{2}{c|}{Kvasir} & \multicolumn{2}{c|}{CVC-ClinicDB} & \multicolumn{2}{c|}{CVC-ColonDB} & \multicolumn{2}{c|}{CVC-T} & \multicolumn{2}{c}{ETIS-Larib}  \\
\cline{2-11}
& mDice & mIOU               & mDice & mIOU                 & mDice & mIOU                & mDice & mIOU                   & mDice & mIOU              \\
\hline
\hline
{{MiT-B3-FaPN}} &  {{0.920}} & {{0.866}} & {0.925} & {0.876} & 0.806 & 0.726 & 0.903 & 0.84  & 0.794 & 0.717 \\

{{MiT-B3-CGNL-FaPN}} &  {{0.923}} & {{0.875}} & {{0.934}} & {{0.888}} & {{0.811}} & {{0.733}} & {0.906} & {0.842}  & {0.817} & {0.738} \\

{{MiT-B3-SE-FaPN}}  & {0.927} & {0.880} & {0.938} & {0.894} & {0.809} & {0.731} & {0.907} & {{0.842}} & {{0.810}} & {{0.730}} \\[2pt]

{{UGCANet}}  & \textbf{0.928} & \textbf{0.881} & \textbf{0.943} & \textbf{0.896} & \textbf{0.827} & \textbf{0.749} & \textbf{0.910} & \textbf{0.847} & \textbf{0.822} & \textbf{0.744} \\[2pt]
\hline
\end{tabular}
}
}
\label{tab:ablation}
\end{table*}

\textbf{Effectiveness of the CGNL}: By utilizing the CGNL module in addition, the ability to connect features by channels in groups of \cite{cgnl} is enhanced. Furthermore, the incorporation of channel attention, as opposed to MiT, yields a slight improvement in Table \ref{tab:ablation}, particularly in the ClinicDB dataset, with an increase of approximately 1\%.

\textbf{Effectiveness of the SE}: When replacing CGNL with SE, the results changed but not significantly. It seems that both modules add a channel attention mechanism to the MiT backbone to increase the representation of features.

\textbf{Combination of CGNL and SE}: After each feature group has learned
internal information, the SE module combines and enhances its channel
information. Our SFMNet performs best when the feature group and SE modules are utilized. Generally, the results of either the upper gastrointestinal dataset or the polyp dataset show a 1-2\% improvement.





\section{Conclusion}
\label{sec:conclusion}
This study introduces a novel model called UGCANet that employs a Unified Global Context-Aware Transformer-based architecture. The proposed model demonstrates exceptional performance in addressing both multi-tasking and single-tasking problems. Our experimental findings indicate that UGCANet achieves state-of-the-art results in multi-tasking and exhibits competitive performance compared to prior segmentation approaches.

In future works, we plan to optimize the utilization of the FSM module to improve the performance of our method. The FSM and SE mechanisms differ only in their usage for the encoder and decoder. By potentially adjusting the FSM module or even the SE module, we can ensure that the model gains more comprehensive information before entering the decoder branch, resulting in enhanced outcomes.



\bibliographystyle{unsrt}
\bibliography{arxiv}

\end{document}